\documentclass[letterpaper,10pt,conference]{ieeeconf}

\IEEEoverridecommandlockouts
\usepackage{amsmath,amssymb}
\usepackage{newtxtext}
\usepackage{booktabs}
\usepackage{array}
\usepackage{graphicx}
\usepackage{xcolor}
\usepackage{url}
\usepackage{cite}
\usepackage{placeins}
\usepackage[colorlinks=true,linkcolor=blue,citecolor=blue,urlcolor=blue]{hyperref}
\let\citeplain\cite
\renewcommand{\cite}[1]{\textcolor{blue}{\citeplain{#1}}}
\graphicspath{{frictionflow_assets/}}

\definecolor{draftred}{RGB}{190,0,0}

\newcommand{\q}{\mathbf{q}}

\newcommand{\uinput}{\mathbf{u}}
\newcommand{\R}{\mathbb{R}}

\title{History-Conditioned Flow Matching for\\
Probabilistic Dynamics of Tendon-Driven Continuum Robots}

\author{Hang Yang, Tingcong Liu, Junjie Xiong, Fangju Yang, and Ke Wu%
}

\begin{document}
\raggedbottom
\maketitle

\begin{abstract}
Deterministic dynamics modeling of tendon-driven continuum robots remains challenging owing to uncertainties in material behavior, tendon transmission, friction, and contact. Measured joint configurations and nominal tendon commands do not fully characterize these internal mechanical factors, leaving uncertainty in the subsequent motion. We therefore develop a history-conditioned, physics-informed flow-matching framework for probabilistic dynamics prediction, using motion and actuation histories to predict the distribution of the next complete joint configuration. By recursively sampling next-step configurations under prescribed commands, the model predicts distributions of future whole-body motions. In simulation, scenario-specific models achieve five-second trajectory Energy Scores (lower is better) of 12.05~mm under internal friction variation and 9.29~mm under unobserved actuation disturbances. Relative to the conditional variational autoencoder and diffusion baselines, Flow attains lower Energy Scores and coverage closer to the nominal level in both scenarios. Ablations support history and structural conditioning in both scenarios. On the physical robot, predictions under two tendon-command profiles excluded from training capture the principal motion sequences, with five-second Energy Scores of 11.91 and 11.42~mm, lower than the compared baselines. The predicted-to-measured spread ratios are 1.65 and 1.22 (closer to 1 is better). These results support history-conditioned probabilistic dynamics prediction under incomplete mechanical observations.
\end{abstract}


\section{Introduction}
\label{sec:introduction}

Deterministic dynamics models underpin motion prediction and model-based control of continuum robots~\cite{armanini2023overview}. In tendon-driven continuum robots, motion depends on structural deformation, tendon force transmission, and frictional contact~\cite{kang2017pneumatic,moses2015cableguide,roy2017friction,kato2015tension}. Identifying the associated material and transmission parameters is challenging~\cite{chen2025review,roy2017friction}, while limited sensing leaves some internal mechanical states unobserved~\cite{russo2023overview}. Uncertainty in these parameters and states can therefore affect the accuracy of motion predictions. A deterministic prediction based on nominal parameters and estimated states does not by itself quantify uncertainty arising from incomplete mechanical information~\cite{kim2021probabilistic}.

\begin{figure*}[!t]
    \centering
    \includegraphics[width=0.98\textwidth]{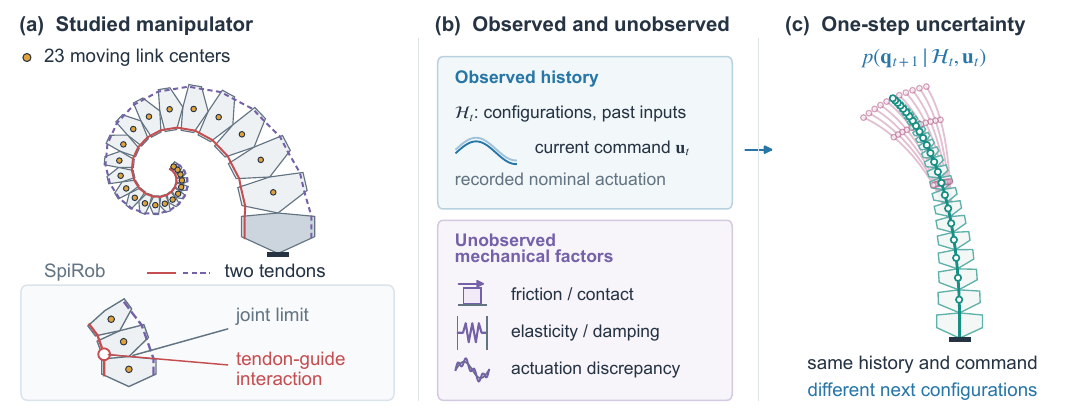}
    \caption{Probabilistic prediction for SpiRob. (a) Two-tendon manipulator with 23 moving-link centers. (b) Observed history and unobserved mechanical factors. (c) Alternative next configurations under the same history and command, with a common base and link geometry.}
    \label{fig:frictionflow_problem}
\end{figure*}

Probabilistic approaches quantify this uncertainty through Bayesian filtering~\cite{kim2021probabilistic}, continuous-time state estimation~\cite{teetaert2025stochastic}, and the propagation of state distributions for control~\cite{honji2025fokkerplanck}. These transition-based formulations commonly adopt a Markov description~\cite{kim2021probabilistic,fox1999markov,honji2025fokkerplanck}, in which the current state and input are assumed to summarize all information needed for the next step. For tendon-driven continuum robots, however, unobserved mechanical states associated with friction, contact, elasticity~\cite{wang2025spirobs}, and tendon-transmission nonlinearities can retain the effects of prior loading~\cite{kato2016hysteresis}. The current measured configuration and nominal command may therefore be insufficient to characterize subsequent motion~\cite{cho2024hysteresis}. This non-Markovian dependence in the observed motion complicates the prediction of conditional motion distributions using model-based probabilistic formulations.

Data-driven generative models offer an alternative by learning configuration distributions directly from recorded motions without an explicit physical model of the unresolved mechanical effects. Previous studies have modeled tendon-conditioned geometry using mixture-density networks~\cite{thompson2024mdn} and multimodal trajectories using conditional variational autoencoders~\cite{ivanovic2020cvae}. Diffusion models~\cite{ho2020ddpm} have also been applied to uncertainty-aware planning~\cite{sun2023plancp}. Conditioning on observation history can help generative models capture dependencies associated with unobserved mechanical states. However, iterative sampling in diffusion models~\cite{ding2025rfmp} can incur substantial computational cost~\cite{ye2024tcfm}, which accumulates during recursive multi-step prediction. Flow matching provides a regression-based formulation for generative modeling~\cite{lipman2023flowmatching}, with rectified flow offering potential for few-step, computationally efficient sampling~\cite{liu2023rectifiedflow}.

These studies highlight three related challenges: nominal deterministic predictions do not by themselves quantify predictive uncertainty from incomplete mechanical information; model-based probabilistic formulations struggle to represent the observed non-Markovian, history-dependent behavior of tendon-driven robots; and recursive distribution prediction requires a balance between predictive quality and sampling cost. To address these challenges, we propose a history-conditioned, physics-informed flow matching framework for probabilistic dynamics prediction of tendon-driven continuum robots. Our main contributions are as follows:
\begin{itemize}
    \widowpenalty=10000
    \item We formulate probabilistic dynamics prediction under partial observability as a distribution over the next joint configuration, conditioned on motion and actuation histories to account for dependencies not captured by instantaneous measurements.
    \item We develop a physics-informed conditional flow matching framework that conditions configuration sampling on a nominal dynamics reference. A kinematics-based distributional loss supervises recursively generated motions.
    \item Simulations and experiments show lower Energy Scores and higher mean-motion prediction accuracy than the compared probabilistic baselines. Ablation studies support the benefits of history and structural conditioning.
\end{itemize}

\section{Problem Formulation}
\label{sec:problem}

\subsection{Studied Manipulator and Available Measurements}

The studied system is SpiRob, a fixed-base planar continuum manipulator actuated by two routed tendons~\cite{wang2025spirobs,wang2024friction}. We represent its articulated body by 23 revolute joints, with relative joint angles $\q_t\in\R^{23}$ as generalized configuration coordinates at sampling index $t$. The recorded input $\uinput_t=[u_{1,t},u_{2,t}]^\top\in\R^2$ contains the two nominal tendon commands applied during the next sampling interval.

Let $\mathbf p_{i,t}\in\R^2$ denote the center of moving link $i$ in fixed-base planar coordinates, and let $\mathbf P_t$ stack all 23 centers, excluding the fixed base. The physical manipulator operates on a horizontal supporting surface; both coordinates lie in this working plane. Known link geometry determines these positions through forward kinematics~\cite{armanini2023overview}:
\begin{equation}
    \mathbf P_t=\mathrm{FK}(\q_t)
    =[\mathbf p_{1,t},\ldots,\mathbf p_{23,t}]^\top
    \in\R^{23\times2},
\label{eq:forward_kinematics}
\end{equation}
Tracked markers provide partial observations of the body geometry. We assume that $\q_t$ is available from measurement or causal kinematic reconstruction. The available configuration and applied-command history, starting at the first recorded observation, is denoted by $\mathcal H_t$:
\begin{equation}
    \mathcal H_t=\{\q_{0:t},\uinput_{0:t-1}\}.
\label{eq:causal_history}
\end{equation}
The current command $\uinput_t$ is specified separately. Configuration history supplies preceding motion information; at initialization, $\mathcal H_0=\{\q_0\}$. Fig.~\ref{fig:frictionflow_problem} illustrates the manipulator and its observed and unobserved quantities.

\subsection{Probabilistic Configuration Dynamics}

Configuration alone does not specify the complete mechanical state of a continuum manipulator~\cite{armanini2023overview}. Let $\boldsymbol\xi_t$ collect the remaining mechanical state, uncertain parameters, and actuation disturbances over the next sampling interval. These quantities account for elastic and damping effects, tendon transmission, friction, contact, and deviations between nominal and actual actuation~\cite{armanini2023overview,moses2015cableguide,roy2017friction}. With these quantities specified, the discrete-time configuration dynamics $F_t$ over one sampling interval take the form
\begin{equation}
    \q_{t+1}=F_t(\q_t,\uinput_t,\boldsymbol\xi_t),
\label{eq:hidden_transition}
\end{equation}
Equation~\eqref{eq:hidden_transition} assumes that $\boldsymbol\xi_t$ contains all remaining information required for the next-step transition. Thus, given $\q_t$, $\uinput_t$, and $\boldsymbol\xi_t$, the next configuration is independent of earlier history. The available history and command condition the distribution of $\boldsymbol\xi_t$~\cite{kaelbling1998planning}. Applying the law of total probability under this conditional-independence assumption gives
\begin{equation}
\begin{split}
p(\q_{t+1}\mid\mathcal H_t,\uinput_t)
=\int &p(\q_{t+1}\mid\q_t,\uinput_t,\boldsymbol\xi_t)\\
&p(\boldsymbol\xi_t\mid\mathcal H_t,\uinput_t)\,d\boldsymbol\xi_t.
\end{split}
\label{eq:conditional_transition}
\end{equation}
Here $p$ denotes conditional probability distributions. For a given configuration and command, \eqref{eq:hidden_transition} determines the next configuration once the hidden quantities are specified. Uncertainty in these quantities therefore induces a distribution over possible next configurations. We learn the resulting configuration distribution directly from recorded motions and commands, jointly predicting all 23 next-step angles. Forward kinematics~\eqref{eq:forward_kinematics} maps the configuration samples to whole-body shapes.

\section{Methodology}
\label{sec:method}

\begin{figure*}[!t]
    \centering
    \includegraphics[width=0.99\textwidth]{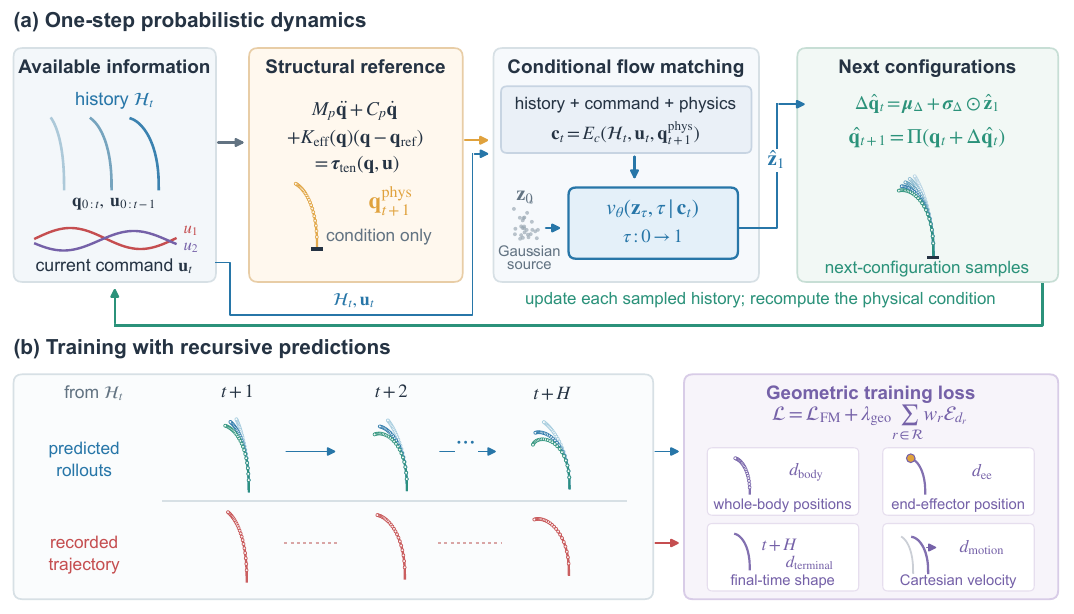}
    \caption{History-conditioned Flow prediction and training. (a) History, command, and a structural reference condition next-configuration samples; each trajectory updates its own history. (b) Geometric Energy Scores supervise recursive motion, illustrated for one history--continuation pair. End-effector position is the last moving-link center.}
    \label{fig:frictionflow_method}
\end{figure*}

We learn a history-conditioned distribution over the robot's next complete joint configuration using conditional flow matching. Motion history supplies information beyond the current configuration and command. A nominal dynamics reference incorporates known mechanical relationships, and kinematics-based losses supervise the resulting motion distributions (Fig.~\ref{fig:frictionflow_method}).

\subsection{Structural Dynamics Conditioning}

The structural reference captures part of the robot's mechanical response, allowing the learned predictor to use known mechanical relationships. The nominal dynamics model~\cite{armanini2023overview,johnson2021firstprinciples} uses inertia, Coriolis--centrifugal, and effective stiffness matrices $M_p,C_p,K_{\mathrm{eff}}\in\R^{23\times23}$, tendon torques $\boldsymbol\tau_{\mathrm{ten}}\in\R^{23}$, and elastic reference configuration $\q_{\mathrm{ref}}\in\R^{23}$. With joint velocity $\dot\q$ and acceleration $\ddot\q$, the nominal balance for horizontal motion, where gravity contributes no generalized joint torque, is
\begin{equation}
\begin{split}
M_p(\q)\ddot\q+C_p(\q,\dot\q)\dot\q\\
+K_{\mathrm{eff}}(\q)(\q-\q_{\mathrm{ref}})
=\boldsymbol\tau_{\mathrm{ten}}(\q,\uinput).
\end{split}
\label{eq:structural_dynamics}
\end{equation}
Here $p$ denotes the nominal model; $K_{\mathrm{eff}}$ includes adjacent-joint coupling and near-limit stiffening.

For ideal frictionless routing, let $\boldsymbol\ell_{\mathrm{ten}}(\q)\in\R^2$ denote tendon lengths, $J_\ell=\partial\boldsymbol\ell_{\mathrm{ten}}/\partial\q\in\R^{2\times23}$ their Jacobian, and $\mathbf T(\uinput)\in\R^2$ the command-dependent nominal tensions. Virtual work gives~\cite{kato2015tension,moses2015cableguide}
\begin{equation}
    \boldsymbol\tau_{\mathrm{ten}}(\q,\uinput)
    =-J_\ell(\q)^\top\mathbf T(\uinput).
\label{eq:tendon_actuation}
\end{equation}
From the joint-limited $\q_t$, backward-difference velocity from the two latest configurations and recorded interval, and command $\uinput_t$, we integrate one sampling interval with joint limits enforced to obtain $\q^{\mathrm{phys}}_{t+1}$. This physical reference enters only as a condition; the Flow learns its relation to the observed transition rather than an imposed additive residual.

\subsection{Conditional Configuration Transition}

A recurrent encoder summarizes the available motion--actuation history in a fixed-dimensional representation. Together with the current command and structural reference, it forms the condition $\mathbf c_t$ through $E_c$:
\begin{equation}
    \mathbf c_t=E_c(\mathcal H_t,\uinput_t,\q^{\mathrm{phys}}_{t+1}).
\label{eq:condition_encoder}
\end{equation}
The history representation provides temporal context for predicting configuration changes influenced by mechanical effects that are not fully captured by instantaneous observations.

Let $\boldsymbol\mu_\Delta,\boldsymbol\sigma_\Delta\in\R^{23}$ be the training-set mean and positive standard deviation of configuration increments. With element-wise division $\oslash$, the Flow target is
\begin{equation}
    \mathbf z_1
    =(\q_{t+1}-\q_t-\boldsymbol\mu_\Delta)
      \oslash\boldsymbol\sigma_\Delta.
\label{eq:normalization}
\end{equation}
For each recorded transition, independently sample $\mathbf z_0\sim\mathcal N(\mathbf0,I_{23})$ and $\tau\sim\mathcal U(0,1)$, where $I_{23}$ is the identity. The linear path $\mathbf z_\tau=(1-\tau)\mathbf z_0+\tau\mathbf z_1$ uses flow time $\tau$, distinct from physical time~\cite{lipman2023flowmatching,liu2023rectifiedflow}. The encoder and velocity field $v_\theta(\cdot)\in\R^{23}$ are jointly trained with parameters $\theta$ by
\begin{equation}
    \mathcal L_{\mathrm{FM}}
    =\frac{1}{23}\mathbb E\!\left[
    \left\|v_\theta(\mathbf z_\tau,\tau\mid\mathbf c_t)
    -(\mathbf z_1-\mathbf z_0)\right\|_2^2\right].
\label{eq:flow_matching_loss}
\end{equation}
The linear interpolation and endpoint-difference target constitute a conditional rectified flow objective with independent source--data coupling~\cite{liu2023rectifiedflow}. Evaluating this one-step objective requires no numerical flow integration.
To generate a sample, integrate the conditional velocity field over flow time:
\begin{equation}
    \frac{d\mathbf z_\tau}{d\tau}
    =v_\theta(\mathbf z_\tau,\tau\mid\mathbf c_t),
    \qquad \tau\in[0,1].
\label{eq:flow_sampling_ode}
\end{equation}
Starting from a Gaussian $\mathbf z_0$, integrating~\eqref{eq:flow_sampling_ode} yields $\hat{\mathbf z}_1$ at $\tau=1$. As shown in Fig.~\ref{fig:frictionflow_method}(a), inverse normalization first recovers the configuration increment:
\begin{equation}
    \Delta\hat\q_t=\boldsymbol\mu_\Delta
    +\boldsymbol\sigma_\Delta\odot\hat{\mathbf z}_1,
\label{eq:sampled_increment}
\end{equation}
where $\odot$ denotes element-wise multiplication. Adding the increment in~\eqref{eq:sampled_increment} to $\q_t$ and projecting onto joint limits gives
\begin{equation}
    \hat\q_{t+1}=\Pi(\q_t+\Delta\hat\q_t),
\label{eq:sampled_transition}
\end{equation}
where $\Pi$ clips each joint angle to $[q_i^{\min},q_i^{\max}]$. Forward kinematics preserves link geometry and connectivity.

\subsection{Geometric Supervision of Recursive Motion}

One-step fitting does not directly supervise errors accumulated under the model's own predictions. We therefore generate $K\geq2$ rollouts of $H$ transitions per recorded history by recursively applying~\eqref{eq:sampled_transition}, each updating its history and physical condition. Known forward kinematics evaluates these predictions in body-position space, avoiding a separate learned shape decoder. Let $\mathbf P=\mathbf P_{t+1:t+H}$ be the measured Cartesian window and $\hat{\mathbf P}^{(k)}$ its $k$th prediction through~\eqref{eq:forward_kinematics}. For a geometric distance $d$, we use the Energy Score (ES) as a distributional training loss~\cite{gneiting2007scoring}:
\begin{equation}
\begin{split}
    \mathcal E_d={}&\frac{1}{K}\sum_{k=1}^K
        d(\hat{\mathbf P}^{(k)},\mathbf P)\\
    &-\frac{1}{2K(K-1)}\sum_{k\ne l}
        d(\hat{\mathbf P}^{(k)},\hat{\mathbf P}^{(l)}).
\end{split}
\label{eq:rollout_energy}
\end{equation}
Scores are averaged over matched history--continuation records (Fig.~\ref{fig:frictionflow_method}(b)).

For a fixed position scale $s_p>0$, the whole-body distance is
\begin{equation}
    d_{\mathrm{body}}(\hat{\mathbf P},\mathbf P)
    =\frac{1}{s_p}\sqrt{\frac{1}{23H}
    \sum_{j=1}^{H}\sum_{i=1}^{23}
    \|\hat{\mathbf p}_{i,t+j}-\mathbf p_{i,t+j}\|_2^2}.
\label{eq:fk_loss}
\end{equation}
The distances $d_{\mathrm{ee}}$ and $d_{\mathrm{terminal}}$ restrict~\eqref{eq:fk_loss} to the last moving-link center or final time, respectively, with corresponding averaging. The motion distance $d_{\mathrm{motion}}$ is the root-mean-square backward-difference velocity error over all centers, scaled by $s_v>0$, using recorded intervals and the observed starting shape for the first difference. For $\mathcal R=\{\mathrm{body},\mathrm{ee},\mathrm{terminal},\mathrm{motion}\}$ and weights $w_r\geq0$, these Energy Score terms form the geometric training loss
\begin{equation}
    \mathcal L_{\mathrm{geo}}=\sum_{r\in\mathcal R}w_r\mathcal E_{d_r}.
\label{eq:geometric_loss}
\end{equation}

\subsection{Training and Recursive Prediction}

\subsubsection{Training}
After one-step training, we progressively extend the rollout window and add the geometric loss in~\eqref{eq:geometric_loss} with weight $\lambda_{\mathrm{geo}}\geq0$:
\begin{equation}
    \mathcal L=\mathcal L_{\mathrm{FM}}
    +\lambda_{\mathrm{geo}}\mathcal L_{\mathrm{geo}}.
\label{eq:complete_training_objective}
\end{equation}
Gradients propagate through sampled transitions and histories, excluding the structural-model branch. Training data determine normalization and geometric scales; $w_r$ are prescribed and $\lambda_{\mathrm{geo}}$ is calibrated using training-set gradients.

\subsubsection{Inference}
All $K$ trajectories start from the same initial history, obtained by averaging the measured histories across repeats, and append their own predicted configurations under prescribed commands. Training rollouts and inference use 40 Euler steps per transition over $\tau\in[0,1]$.

\begin{table}[!t]
\centering
\caption{Evaluation metrics and units ($\downarrow$: lower is better; $\to$: closer is better).}
\label{tab:metric_definitions}
\resizebox{\columnwidth}{!}{%
\begin{minipage}{1.1765\columnwidth}
\normalsize
\setlength{\tabcolsep}{3pt}
\renewcommand{\arraystretch}{1.05}
\setlength{\aboverulesep}{1.4pt}
\setlength{\belowrulesep}{1.4pt}
\begin{tabular}{>{\raggedright\arraybackslash}p{2.05cm}>{\raggedright\arraybackslash}p{\dimexpr\columnwidth-2.05cm-0.75cm-6\tabcolsep\relax}>{\centering\arraybackslash}p{0.75cm}}
\toprule
Metric & Definition & Unit\\
\midrule
\multicolumn{3}{l}{\emph{Mean-motion accuracy (RMSE)}}\\
Shape $\downarrow$ & $\textstyle \sqrt{\frac{1}{JH}\sum_{i,j}\|\overline{\hat{\mathbf p}}_{i,t+j}-\overline{\mathbf p}_{i,t+j}\|_2^2}$ & mm\\[2pt]
EE $\downarrow$ & $\textstyle \sqrt{\frac{1}{H}\sum_j\|\overline{\hat{\mathbf p}}_{J,t+j}-\overline{\mathbf p}_{J,t+j}\|_2^2}$ & mm\\
\midrule
\multicolumn{3}{l}{\emph{Distribution agreement}}\\
ES $\downarrow$ & $\textstyle \frac{1}{N}\sum_n\mathcal E_d^{(n)}$, with~\eqref{eq:rollout_energy} evaluated against repeat~$n$~\cite{gneiting2007scoring,shahroudi2024energyscore}. & mm\\[2pt]
$D\to1$ & $\displaystyle \frac{\frac{1}{K(K-1)}\sum_{k\ne l}d(\hat{\mathbf P}^{(k)},\hat{\mathbf P}^{(l)})}{\frac{1}{N(N-1)}\sum_{n\ne m}d(\mathbf P^{(n)},\mathbf P^{(m)})}$ & --\\
\midrule
\multicolumn{3}{l}{\emph{Pointwise coverage: intervals ($M_{90}$), ellipses ($S_{90}$)}}\\
$M_{90}\to90\%$ & $\textstyle \frac{100}{2NJH}\sum_{n,i,j,c}\mathbf 1\!\left[Q_{0.05}^{ijc}\leq p_{i,t+j,c}^{(n)}\leq Q_{0.95}^{ijc}\right]$ & \%\\[2pt]
$S_{90}\to90\%$ & $\textstyle \frac{100}{NJH}\sum_{n,i,j}\mathbf 1\!\left[(\mathbf e_{ij}^{(n)})^\top\Sigma_{ij}^{-1}\mathbf e_{ij}^{(n)}\leq4.605\right]$ & \%\\
\bottomrule
\end{tabular}
\end{minipage}%
}
\par\vspace{2pt}
\begin{minipage}{\columnwidth}
\footnotesize
$N$: measured repeats; $J=23$, $c=1,2$. Bars: ensemble means; $Q_\alpha$: predicted quantiles; $\mathbf1$: indicator. Pointwise, $\mathbf e=\mathbf p-\overline{\hat{\mathbf p}}$ and $\Sigma=\operatorname{Cov}(\hat{\mathbf p})+(0.5\,\mathrm{mm})^2I_2$, using divisor $K-1$ and a Gaussian reference for $S_{90}$.
\end{minipage}
\vspace{-20pt}
\end{table}

\begin{table}[!b]
\centering
\caption{One-step prediction from measured histories.}
\label{tab:one_step_comparison}
\small
\setlength{\tabcolsep}{4pt}
\begin{tabular*}{\columnwidth}{@{\extracolsep{\fill}}lrrrr@{}}
\toprule
 & \multicolumn{2}{c}{Internal friction variation} & \multicolumn{2}{c}{Actuation disturbances}\\
\cmidrule(lr){2-3}\cmidrule(lr){4-5}
Model & Shape & ES & Shape & ES\\
\midrule
CVAE & 0.645 & 0.454 & 0.594 & 0.459\\
Diffusion & 0.461 & 0.327 & 0.161 & 0.111\\
\textbf{Flow} & \textbf{0.449} & \textbf{0.318} & \textbf{0.103} & \textbf{0.068}\\
\bottomrule
\end{tabular*}
\end{table}

\begin{table*}[!t]
\centering
\caption{Five-second recursive prediction.}
\label{tab:recursive_comparison}
\small
\setlength{\tabcolsep}{8pt}
\begin{tabular*}{\textwidth}{@{\extracolsep{\fill}}llrrrrrr@{}}
\toprule
Scenario & Model & Shape $\downarrow$ & EE $\downarrow$ & ES $\downarrow$ & $D\to1$ & $M_{90}\to90\%$ & $S_{90}\to90\%$\\
\midrule
Internal friction & CVAE & 19.244 & 28.445 & 15.611 & 1.884 & 67.0 & 68.6\\
variation & Diffusion & 14.340 & \textbf{25.296} & 12.407 & \textbf{1.505} & 73.4 & 74.1\\
 & \textbf{Flow} & \textbf{13.917} & 25.469 & \textbf{12.048} & 1.559 & \textbf{80.3} & \textbf{85.3}\\
\midrule
Actuation & CVAE & 15.246 & 22.177 & 15.759 & 0.375 & 20.9 & 28.1\\
disturbances & Diffusion & 10.251 & 14.045 & 10.377 & \textbf{0.896} & 65.9 & 69.6\\
 & \textbf{Flow} & \textbf{8.785} & \textbf{13.438} & \textbf{9.290} & 1.119 & \textbf{70.3} & \textbf{79.0}\\
\bottomrule
\end{tabular*}
\end{table*}

\section{Simulation}
\label{sec:simulation}

To assess prediction of motion distributions under unobserved mechanics, we compare the proposed model with probabilistic baselines under internal friction variation and actuation disturbances. Retrained ablations examine the contributions of history and structural conditioning.

\subsection{Simulation Setup}

\subsubsection{Simulation Scenarios}
We simulate the 23-joint, two-tendon SpiRob in MuJoCo~\cite{todorov2012mujoco}. Predictors receive recorded configurations and nominal commands; mechanical variations and actual disturbances remain unobserved. The two scenarios are defined as follows:

\begin{itemize}
    \item \textbf{Internal friction variation:} Tendon friction loss, joint Coulomb friction, and viscous damping vary across repetitions and links but remain fixed within each execution. Two command sequences follow a common curling procedure.
    \item \textbf{Actuation disturbances:} With fixed mechanics and curled initial configuration, both tendon commands receive small, temporally correlated disturbances generated from Gaussian noise.
\end{itemize}

\subsubsection{Training and Baselines}
Each scenario provides 64 training executions, from which history--transition pairs are extracted. Complete executions remain in separate training, development, and test sets. Independent test repeats use the same nominal commands.

% Draft unified comparison protocol: existing results await replacement by the user's experiments; no experiments were run for this text revision.
Baselines comprise a conditional variational autoencoder (CVAE)~\cite{ivanovic2020cvae} and conditional diffusion~\cite{ho2020ddpm}. In both scenarios, all predictors use causal history to predict configuration increments and have approximately 2.2M parameters. Flow and diffusion use 40 Euler and DDIM steps, respectively.

\subsubsection{Prediction Protocol}
% Draft protocol revision only: existing numerical results have not been recomputed using the mean initial history.
For each command, we initialize 64 sampled trajectories from a common 0.5-s history, obtained by averaging the measured histories across test repeats at each time step. We evaluate the first predicted step and the full 300-transition horizon (5~s at approximately 60~Hz). Recursive predictions use prescribed nominal commands and their own predicted histories. Models use the same test population and mean initial history, and enforce joint limits. Results concern repetitions of known commands, with one training seed per model configuration.

\subsubsection{Evaluation Metrics}
Table~\ref{tab:metric_definitions} groups the measures of mean-motion accuracy, distribution agreement, and coverage. Distances use the RMS body-position distance $d=s_p d_{\mathrm{body}}$ from~\eqref{eq:fk_loss}, expressed in millimeters, with $H=1$ for one-step queries and $H=300$ for recursive forecasts.

\subsection{Results}

\subsubsection{Baseline Comparisons}
\begin{figure}[t]
\centering
\includegraphics[width=\columnwidth]{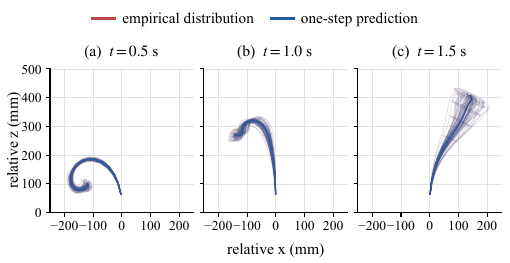}
\caption{Empirical and Flow-predicted one-step configuration distributions under internal friction variation.}
\label{fig:one_step_distribution}
\end{figure}

\begin{figure*}[t]
\centering
\includegraphics[width=\textwidth]{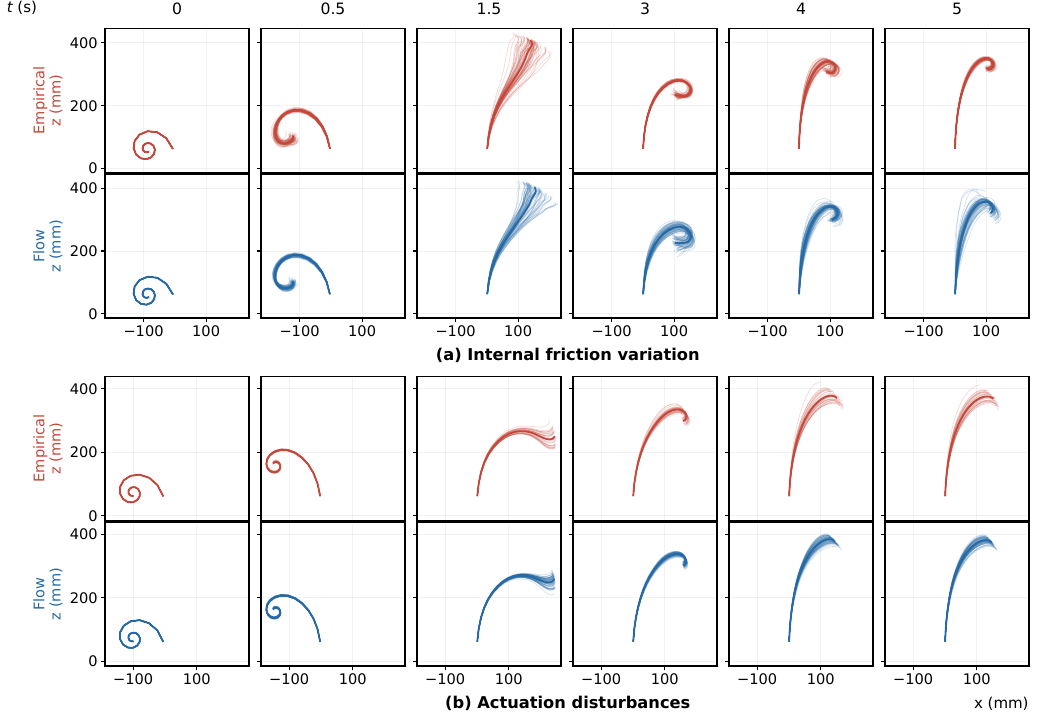}
\caption{Empirical and Flow-predicted motion distributions under (a) internal friction variation and (b) actuation disturbances.}
\label{fig:recursive_distributions}
\end{figure*}

% Discussion follows the currently retained tables; results under the revised unified protocol await the user's experiments.
Under internal friction variation, Flow achieves the lowest one-step mean-shape error and ES among the probabilistic baselines, with an ES of 0.318~mm (Table~\ref{tab:one_step_comparison}). Fig.~\ref{fig:one_step_distribution} illustrates independent next-step distributions at three query times. Over five seconds, Flow's ES is 12.048~mm, compared with 12.407~mm for diffusion and 15.611~mm for CVAE (Table~\ref{tab:recursive_comparison}). Although diffusion has a slightly lower end-effector error and a spread ratio closer to one, Flow provides more accurate mean shapes and spatial coverage closer to the nominal 90\% (85.3\%). Fig.~\ref{fig:recursive_distributions}(a) illustrates the predicted motion distribution under the chirp command.

Under actuation disturbances, Flow achieves the lowest one-step mean-shape error and ES, at 0.103 and 0.068~mm, respectively (Table~\ref{tab:one_step_comparison}). Over five seconds, Flow attains an ES of 9.290~mm, versus 10.377~mm for diffusion and 15.759~mm for CVAE (Table~\ref{tab:recursive_comparison}). Its spatial coverage is 79.0\%, compared with 69.6\% and 28.1\%, respectively. Fig.~\ref{fig:recursive_distributions}(b) shows the corresponding motion sequence and predicted variability.

\setcounter{dbltopnumber}{2}

\subsubsection{History and Structural Conditioning}
Table~\ref{tab:component_comparison} supports both history and structural conditioning. Shortening history to two observations increases ES from 12.048 to 30.707~mm under friction variation and from 9.290 to 57.666~mm under actuation disturbances; removing the structural reference raises ES to 14.968 and 30.920~mm. The short-history friction model achieves 90.5\% spatial coverage but excessive dispersion ($D=4.100$). Full conditioning combines lower ES with spread closer to the measured variability ($D=1.559$).

\begin{table}[!htbp]
\centering
\caption{Five-second conditioning ablations.}
\label{tab:component_comparison}
\small
\setlength{\tabcolsep}{4pt}
\begin{tabular*}{\columnwidth}{@{\extracolsep{\fill}}lrrrr@{}}
\toprule
Variant & Shape & ES & $D$ & $S_{90}$\\
\midrule
\multicolumn{5}{l}{\emph{Internal friction variation}}\\
Short history & 43.305 & 30.707 & 4.100 & \textbf{90.5}\\
Without physics & 17.827 & 14.968 & 1.697 & 57.8\\
\textbf{Full conditioning} & \textbf{13.917} & \textbf{12.048} & \textbf{1.559} & 85.3\\
\midrule
\multicolumn{5}{l}{\emph{Actuation disturbances}}\\
Short history & 72.366 & 57.666 & 3.405 & 36.5\\
Without physics & 36.665 & 30.920 & 1.300 & 22.5\\
\textbf{Full conditioning} & \textbf{8.785} & \textbf{9.290} & \textbf{1.119} & \textbf{79.0}\\
\bottomrule
\end{tabular*}
\end{table}

\begin{figure}[!b]
\centering
\includegraphics[width=\columnwidth]{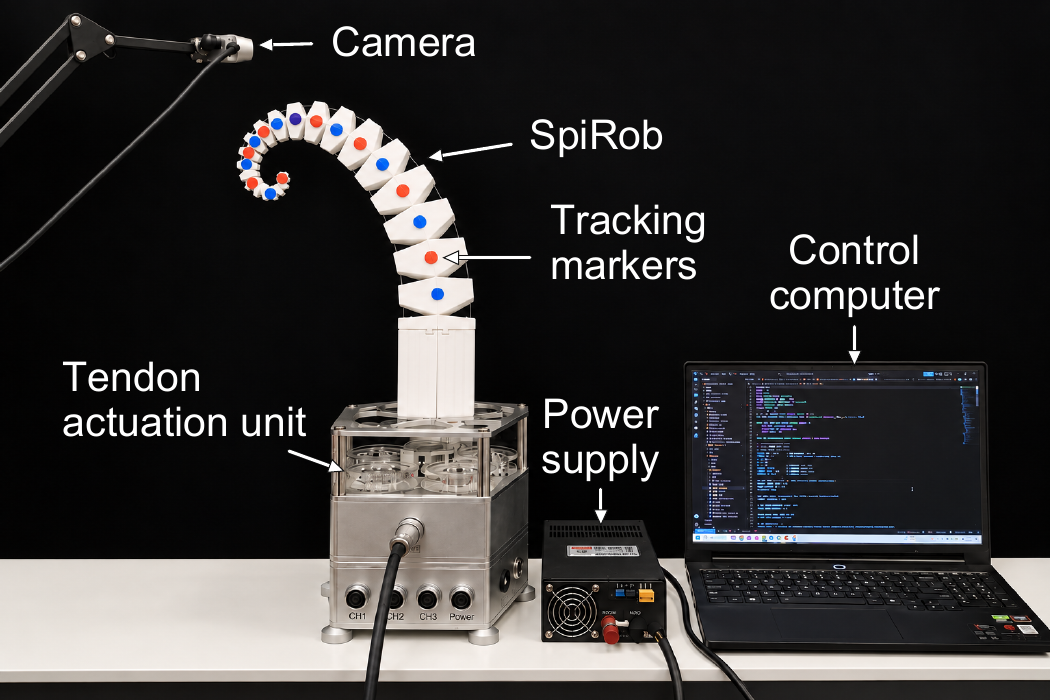}
\caption{Experimental setup (background enhanced for clarity).}
\label{fig:physical_setup}
\end{figure}

\begin{table}[!htbp]
\centering
\caption{Model size and computational cost.}
\label{tab:efficiency}
\small
\setlength{\tabcolsep}{3pt}
\begin{tabular*}{\columnwidth}{@{\extracolsep{\fill}}lrrrr@{}}
\toprule
Model & Params (M) & Training & Next step & Five seconds\\
 & & (min) & (ms) & (s)\\
\midrule
CVAE & $\approx 2.2$ & \textbf{30.33} & \textbf{10.199} & \textbf{0.390}\\
Diffusion & $\approx 2.2$ & 32.37 & 29.670 & 8.250\\
\textbf{Flow} & $\approx 2.2$ & 30.94 & 21.600 & 4.940\\
\bottomrule
\end{tabular*}
\end{table}

\begingroup\widowpenalty=10000
\subsubsection{Computational Cost}

Table~\ref{tab:efficiency} summarizes training and inference costs under internal friction variation on an Apple M4 Max. Including conditioning, sampling, and forward kinematics, Flow generates 64 next-step states in 21.6~ms and propagates 64 trajectories over 300 steps (5~s) in 4.94~s. These times are 27.2\% and 40.1\% lower than those of diffusion, respectively. Although CVAE remains faster, Flow's 21.6-ms next-step latency suggests potential for online prediction in lower-frequency control loops.
\par\endgroup

\section{Experiments}
\label{sec:physical}

\subsection{Experimental Setup}

SpiRob operates on a horizontal surface and is driven by two torque-controlled J60-series tendon actuators (DEEP Robotics). A fixed camera tracks 18 body-mounted markers (Fig.~\ref{fig:physical_setup}). We collect repeated motion recordings under five different two-channel tendon-command profiles, starting from a curled reset target. Recordings are aligned at 20~Hz after tendon tightening. Reconstructed configuration increments and reliable marker measurements provide training supervision. The structural reference combines nominal inertial properties with stiffness, damping, and effective actuation parameters fitted using training data.

\begin{figure*}[t]
\centering
\includegraphics[width=\textwidth]{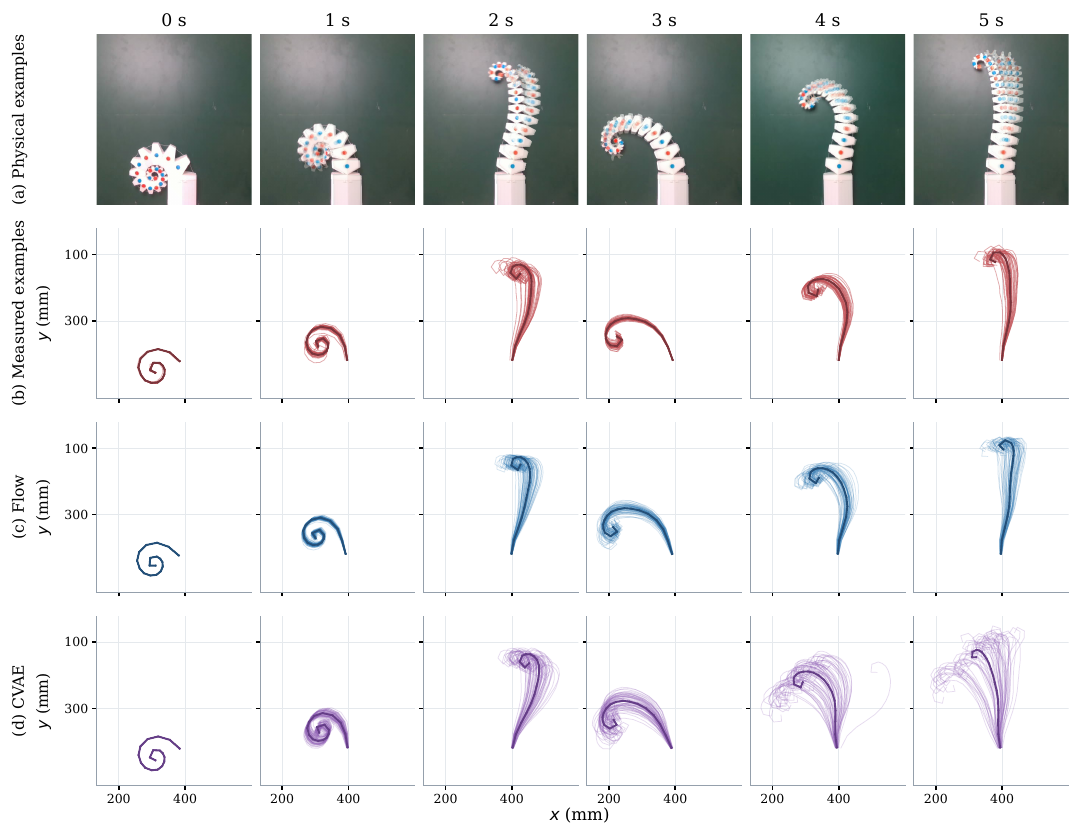}
\caption{Physical motion under Test input 2: (a) robot photographs, (b) measured motion distributions, (c) Flow predictions, and (d) CVAE predictions.}
\label{fig:physical_distributions}
\end{figure*}

\subsection{Experimental Protocol}

% Draft unified comparison protocol as above; existing physical results have not been recomputed.
The data are split by complete executions: 80 training and 20 validation repeats under three input profiles, and 54 test repeats under two held-out profiles, denoted Test input 1 and Test input 2. All predictors are trained exclusively on real recordings and evaluated on the same test set. Baselines, model sizes, prediction targets, and solver step counts follow the simulation comparison.

For each test input, we generate 64 trajectories from a shared 0.5-s measured history. Each trajectory advances for 100 steps (5~s) under the prescribed commands, updating its history with its own predictions. At each transition, Flow independently rejection-samples standard Gaussian source coordinates within $[-1,1]$; the probabilistic baselines retain untruncated Gaussian sources.

We evaluate the metrics in Table~\ref{tab:metric_definitions} using valid marker observations. The same predicted ensemble is compared with each measured repeat; RMSE measures the error of its mean, and RMSE, ES, and coverage are averaged equally over repeats. The spread ratio $D$ uses the common valid support across repeats. Distances use the nominal scale of 1.1119~mm/pixel, with $(0.556~\mathrm{mm})^2I_2$ covariance regularization for spatial coverage.

\subsection{Results}

Fig.~\ref{fig:physical_distributions} shows that Flow follows the principal curling and extension sequence under Test input 2. In Table~\ref{tab:physical_comparison}, Flow achieves the lowest marker RMSE and ES under both test inputs: RMSEs of 16.29 and 15.86~mm, and ES values of 11.91 and 11.42~mm.

CVAE's coverage is closer to the nominal 90\%, but is accompanied by spread ratios of 4.717 and 3.780, indicating substantially broader predictive ensembles than the measured repeats. Flow's corresponding ratios, 1.648 and 1.217, are closer to one. Thus, despite lower pointwise coverage, Flow combines more accurate motion prediction with a closer match to empirical spread.

\begin{table}[!htbp]
\centering
\caption{Physical motion prediction.}
\label{tab:physical_comparison}
\small
\setlength{\tabcolsep}{3pt}
\begin{tabular*}{\columnwidth}{@{\extracolsep{\fill}}llrrrrr@{}}
\toprule
Input & Model & RMSE & ES & $D$ & $M_{90}$ & $S_{90}$\\
\midrule
1 & CVAE & 23.10 & 15.45 & 4.717 & \textbf{82.1} & \textbf{79.5}\\
 & Diffusion & 26.79 & 20.81 & 1.920 & 57.7 & 50.3\\
 & \textbf{Flow} & \textbf{16.29} & \textbf{11.91} & \textbf{1.648} & 50.8 & 45.6\\
\midrule
2 & CVAE & 38.95 & 25.92 & 3.780 & \textbf{82.3} & \textbf{86.5}\\
 & Diffusion & 40.48 & 31.58 & 1.775 & 54.6 & 47.9\\
 & \textbf{Flow} & \textbf{15.86} & \textbf{11.42} & \textbf{1.217} & 64.9 & 62.9\\
\bottomrule
\end{tabular*}
\end{table}

\section{Conclusion}

We presented a history-conditioned, physics-informed flow matching framework for probabilistic dynamics prediction of a tendon-driven continuum robot. In both simulated scenarios, Flow achieves lower Energy Scores and coverage closer to the nominal level than the compared baselines; ablations support history and structural conditioning. Under two held-out physical input profiles, Flow yields lower prediction errors and spread ratios closer to one, despite lower pointwise coverage than CVAE. These results support combining motion history with mechanical and geometric structure for probabilistic prediction under incomplete observations. Future work will explore the predicted motion distributions for uncertainty-aware control and contact-rich manipulation.

\section*{Acknowledgments}

The authors used ChatGPT/Codex (OpenAI) for code and manuscript assistance, and an AI tool to enhance the background of Fig.~\ref{fig:physical_setup}. The authors remain responsible for all technical content and results.

\bibliographystyle{IEEEtran}
\bibliography{references_frictionflow}

\begin{thebibliography}{10}
\providecommand{\url}[1]{#1}
\csname url@rmstyle\endcsname
\providecommand{\newblock}{\relax}
\providecommand{\bibinfo}[2]{#2}
\providecommand\BIBentrySTDinterwordspacing{\spaceskip=0pt\relax}
\providecommand\BIBentryALTinterwordstretchfactor{4}
\providecommand\BIBentryALTinterwordspacing{\spaceskip=\fontdimen2\font plus
\BIBentryALTinterwordstretchfactor\fontdimen3\font minus \fontdimen4\font\relax}
\providecommand\BIBforeignlanguage[2]{{%
\expandafter\ifx\csname l@#1\endcsname\relax
\typeout{** WARNING: IEEEtran.bst: No hyphenation pattern has been}%
\typeout{** loaded for the language `#1'. Using the pattern for}%
\typeout{** the default language instead.}%
\else
\language=\csname l@#1\endcsname
\fi
#2}}

\bibitem{armanini2023overview}
C.~Armanini, F.~Boyer, A.~T. Mathew, C.~Duriez, and F.~Renda, ``Soft robots modeling: A structured overview,'' \emph{IEEE Transactions on Robotics}, vol.~39, no.~3, pp. 1728--1748, 2023.

\bibitem{kang2017pneumatic}
R.~Kang, Y.~Guo, L.~Chen, D.~T. Branson, and J.~S. Dai, ``Design of a pneumatic muscle based continuum robot with embedded tendons,'' \emph{IEEE/ASME Transactions on Mechatronics}, vol.~22, no.~2, pp. 751--761, 2017.

\bibitem{moses2015cableguide}
M.~S. Moses, R.~J. Murphy, M.~D.~M. Kutzer, and M.~Armand, ``Modeling cable and guide channel interaction in a high-strength cable-driven continuum manipulator,'' \emph{IEEE/ASME Transactions on Mechatronics}, vol.~20, no.~6, pp. 2876--2889, 2015.

\bibitem{roy2017friction}
R.~Roy, L.~Wang, and N.~Simaan, ``Modeling and estimation of friction, extension, and coupling effects in multisegment continuum robots,'' \emph{IEEE/ASME Transactions on Mechatronics}, vol.~22, no.~2, pp. 909--920, 2017.

\bibitem{kato2015tension}
T.~Kato, I.~Okumura, S.-E. Song, A.~J. Golby, and N.~Hata, ``Tendon-driven continuum robot for endoscopic surgery: Preclinical development and validation of a tension propagation model,'' \emph{IEEE/ASME Transactions on Mechatronics}, vol.~20, no.~5, pp. 2252--2263, 2015.

\bibitem{chen2025review}
Z.~Chen, F.~Renda, A.~Le~Gall, L.~Mocellin, M.~Bernabei, T.~Dangel, G.~Ciuti, M.~Cianchetti, and C.~Stefanini, ``Data-driven methods applied to soft robot modeling and control: A review,'' \emph{IEEE Transactions on Automation Science and Engineering}, vol.~22, pp. 2241--2256, 2025.

\bibitem{russo2023overview}
M.~Russo, S.~M.~H. Sadati, X.~Dong, A.~Mohammad, I.~D. Walker, C.~Bergeles, K.~Xu, and D.~A. Axinte, ``Continuum robots: An overview,'' \emph{Advanced Intelligent Systems}, vol.~5, no.~5, p. 2200367, 2023.

\bibitem{kim2021probabilistic}
D.~Kim, M.~Park, and Y.-L. Park, ``Probabilistic modeling and bayesian filtering for improved state estimation for soft robots,'' \emph{IEEE Transactions on Robotics}, vol.~37, no.~5, pp. 1728--1741, 2021.

\bibitem{teetaert2025stochastic}
S.~Teetaert, S.~Lilge, J.~Burgner-Kahrs, and T.~D. Barfoot, ``A stochastic framework for continuous-time state estimation of continuum robots,'' \emph{arXiv preprint arXiv:2510.01381}, 2025.

\bibitem{honji2025fokkerplanck}
S.~Honji and T.~Wada, ``Model predictive control for a soft robotic finger with stochastic behavior based on fokker--planck equation,'' in \emph{2025 IEEE 8th International Conference on Soft Robotics (RoboSoft)}, 2025, pp. 1--6.

\bibitem{fox1999markov}
D.~Fox, W.~Burgard, and S.~Thrun, ``Markov localization for mobile robots in dynamic environments,'' \emph{Journal of Artificial Intelligence Research}, vol.~11, pp. 391--427, 1999.

\bibitem{wang2025spirobs}
Z.~Wang, N.~M. Freris, and X.~Wei, ``{SpiRobs}: Logarithmic spiral-shaped robots for versatile grasping across scales,'' \emph{Device}, vol.~3, no.~4, p. 100646, 2025.

\bibitem{kato2016hysteresis}
T.~Kato, I.~Okumura, H.~Kose, K.~Takagi, and N.~Hata, ``Tendon-driven continuum robot for neuroendoscopy: Validation of extended kinematic mapping for hysteresis operation,'' \emph{International Journal of Computer Assisted Radiology and Surgery}, vol.~11, no.~4, pp. 589--602, 2016.

\bibitem{cho2024hysteresis}
B.~Y. Cho, D.~S. Esser, J.~Thompson, B.~Thach, R.~J. Webster, and A.~Kuntz, ``Accounting for hysteresis in the forward kinematics of nonlinearly-routed tendon-driven continuum robots via a learned deep decoder network,'' \emph{IEEE Robotics and Automation Letters}, vol.~9, no.~11, pp. 9263--9270, 2024.

\bibitem{thompson2024mdn}
J.~Thompson, B.~Y. Cho, D.~S. Brown, and A.~Kuntz, ``Modeling kinematic uncertainty of tendon-driven continuum robots via mixture density networks,'' in \emph{2024 International Symposium on Medical Robotics}, 2024, pp. 1--7.

\bibitem{ivanovic2020cvae}
B.~Ivanovic, K.~Leung, E.~Schmerling, and M.~Pavone, ``Multimodal deep generative models for trajectory prediction: A conditional variational autoencoder approach,'' \emph{IEEE Robotics and Automation Letters}, vol.~6, no.~2, pp. 295--302, 2021.

\bibitem{ho2020ddpm}
J.~Ho, A.~Jain, and P.~Abbeel, ``Denoising diffusion probabilistic models,'' in \emph{Advances in Neural Information Processing Systems}, vol.~33, 2020, pp. 6840--6851.

\bibitem{sun2023plancp}
J.~Sun, Y.~Jiang, J.~Qiu, P.~Nobel, M.~J. Kochenderfer, and M.~Schwager, ``Conformal prediction for uncertainty-aware planning with diffusion dynamics model,'' in \emph{Advances in Neural Information Processing Systems}, vol.~36, 2023, pp. 80\,324--80\,337.

\bibitem{ding2025rfmp}
H.~Ding, N.~Jaquier, J.~Peters, and L.~Rozo, ``Fast and robust visuomotor riemannian flow matching policy,'' \emph{IEEE Transactions on Robotics}, vol.~41, pp. 5327--5343, 2025.

\bibitem{ye2024tcfm}
S.~Ye and M.~C. Gombolay, ``Efficient trajectory forecasting and generation with conditional flow matching,'' in \emph{2024 IEEE/RSJ International Conference on Intelligent Robots and Systems (IROS)}, 2024, pp. 2816--2823.

\bibitem{lipman2023flowmatching}
Y.~Lipman, R.~T.~Q. Chen, H.~Ben-Hamu, M.~Nickel, and M.~Le, ``Flow matching for generative modeling,'' in \emph{The Eleventh International Conference on Learning Representations}, 2023.

\bibitem{liu2023rectifiedflow}
X.~Liu, C.~Gong, and Q.~Liu, ``Flow straight and fast: Learning to generate and transfer data with rectified flow,'' in \emph{The Eleventh International Conference on Learning Representations}, 2023.

\bibitem{wang2024friction}
Z.~Wang and N.~M. Freris, ``Exploiting frictional effects to reproduce octopus-like reaching movements with a cable-driven spiral robot,'' in \emph{2024 IEEE 7th International Conference on Soft Robotics (RoboSoft)}, 2024, pp. 537--542.

\bibitem{kaelbling1998planning}
L.~P. Kaelbling, M.~L. Littman, and A.~R. Cassandra, ``Planning and acting in partially observable stochastic domains,'' \emph{Artificial Intelligence}, vol. 101, no. 1--2, pp. 99--134, 1998.

\bibitem{johnson2021firstprinciples}
C.~C. Johnson, T.~Quackenbush, T.~Sorensen, D.~Wingate, and M.~D. Killpack, ``Using first principles for deep learning and model-based control of soft robots,'' \emph{Frontiers in Robotics and AI}, vol.~8, p. 654398, 2021.

\bibitem{gneiting2007scoring}
T.~Gneiting and A.~E. Raftery, ``Strictly proper scoring rules, prediction, and estimation,'' \emph{Journal of the American Statistical Association}, vol. 102, no. 477, pp. 359--378, 2007.

\bibitem{shahroudi2024energyscore}
N.~Shahroudi, M.~Lepson, and M.~Kull, ``Evaluation of trajectory distribution predictions with energy score,'' in \emph{Proceedings of the 41st International Conference on Machine Learning}, ser. Proceedings of Machine Learning Research, vol. 235, 2024, pp. 44\,322--44\,341.

\bibitem{todorov2012mujoco}
E.~Todorov, T.~Erez, and Y.~Tassa, ``{MuJoCo}: A physics engine for model-based control,'' in \emph{2012 IEEE/RSJ International Conference on Intelligent Robots and Systems}, 2012, pp. 5026--5033.

\end{thebibliography}

\end{document}